\documentclass[conference]{IEEEtran}  

\IEEEoverridecommandlockouts                              

\usepackage{cite}
\usepackage{amsmath,amssymb,amsfonts}
\usepackage{algorithmic}
\usepackage{graphicx}
\usepackage{textcomp}
\usepackage{xcolor}
\usepackage{pifont}
\newcommand{\cmark}{\textcolor{green}{\ding{52}}}
\newcommand{\xmark}{\textcolor{red}{\ding{55}}}
\def\BibTeX{{\rm B\kern-.05em{\sc i\kern-.025em b}\kern-.08em
    T\kern-.1667em\lower.7ex\hbox{E}\kern-.125emX}}                                  

\usepackage{graphicx} 
\usepackage{amsmath} 
\usepackage{amssymb}  

\usepackage[caption=false]{subfig}
\usepackage{tabularx}
\usepackage{wasysym}
\usepackage{multirow}
\usepackage{booktabs}
\usepackage{comment}
\usepackage{xcolor}

\usepackage{graphicx} 
\usepackage{subfig}   
\title{\LARGE \bf
Whole-Body Proprioceptive Morphing:
\\ A Modular Soft Gripper for Robust Cross-Scale Grasping
}
\author{
Dong Heon Han$^{1}$, Xiaohao Xu$^{2,\dagger}$, Yuxi Chen$^{1}$, Yusheng Zhou$^{2}$, Xinqi Zhang$^{1}$, \\ Jiaqi Wang$^{2}$, Daniel Bruder$^{1}$, Xiaonan Huang$^{2,*}  $
\thanks{$^{1}$The authors are with the  Mechanical Engineering Department at the University of Michigan-Ann Arbor, Ann Arbor, MI, USA
        {\tt\small \{dongheon,ethansab,dadaaa,dbruder\}@umich.edu}}%
 \thanks{$^{2}$The authors are with the Robotics Department at the University of Michigan-Ann Arbor,  Ann Arbor, MI, USA  {\tt\small \{xiaohaox, yszhou, wangjq, xiaonanh\}@umich.edu}} \thanks{$\dagger$ Project Lead \quad $*$ Correspondence Author }%
 }

\begin{document}

\maketitle
\thispagestyle{empty}
\pagestyle{empty}


\begin{abstract}
Biological systems, such as the octopus, exhibit masterful cross-scale manipulation by adaptively reconfiguring their entire form, a capability that remains elusive in robotics. Conventional soft grippers, while compliant, are mostly constrained by a fixed global morphology, and prior shape-morphing efforts have been largely confined to localized deformations, failing to replicate this biological dexterity. Inspired by this natural exemplar, we introduce the paradigm of collaborative, whole-body proprioceptive morphing, realized in a modular soft gripper architecture. Our design is a distributed network of modular self-sensing pneumatic actuators that enables the gripper to intelligently reconfigure its entire topology, achieving multiple morphing states that are controllable to form diverse polygonal shapes. By integrating rich proprioceptive feedback from embedded sensors, our system can seamlessly transition from a precise pinch to a large envelope grasp. We experimentally demonstrate that this approach expands the grasping envelope and enhances generalization across diverse object geometries (standard and irregular) and scales (up to 10$\times$), while also unlocking novel manipulation modalities such as multi-object and internal hook grasping. This work presents a low-cost, easy-to-fabricate, and scalable framework that fuses distributed actuation with integrated sensing, offering a new pathway toward achieving biological levels of dexterity in robotic manipulation.
\end{abstract}



\section{Introduction}

\begin{figure}
    \centering    \includegraphics[width=\linewidth]{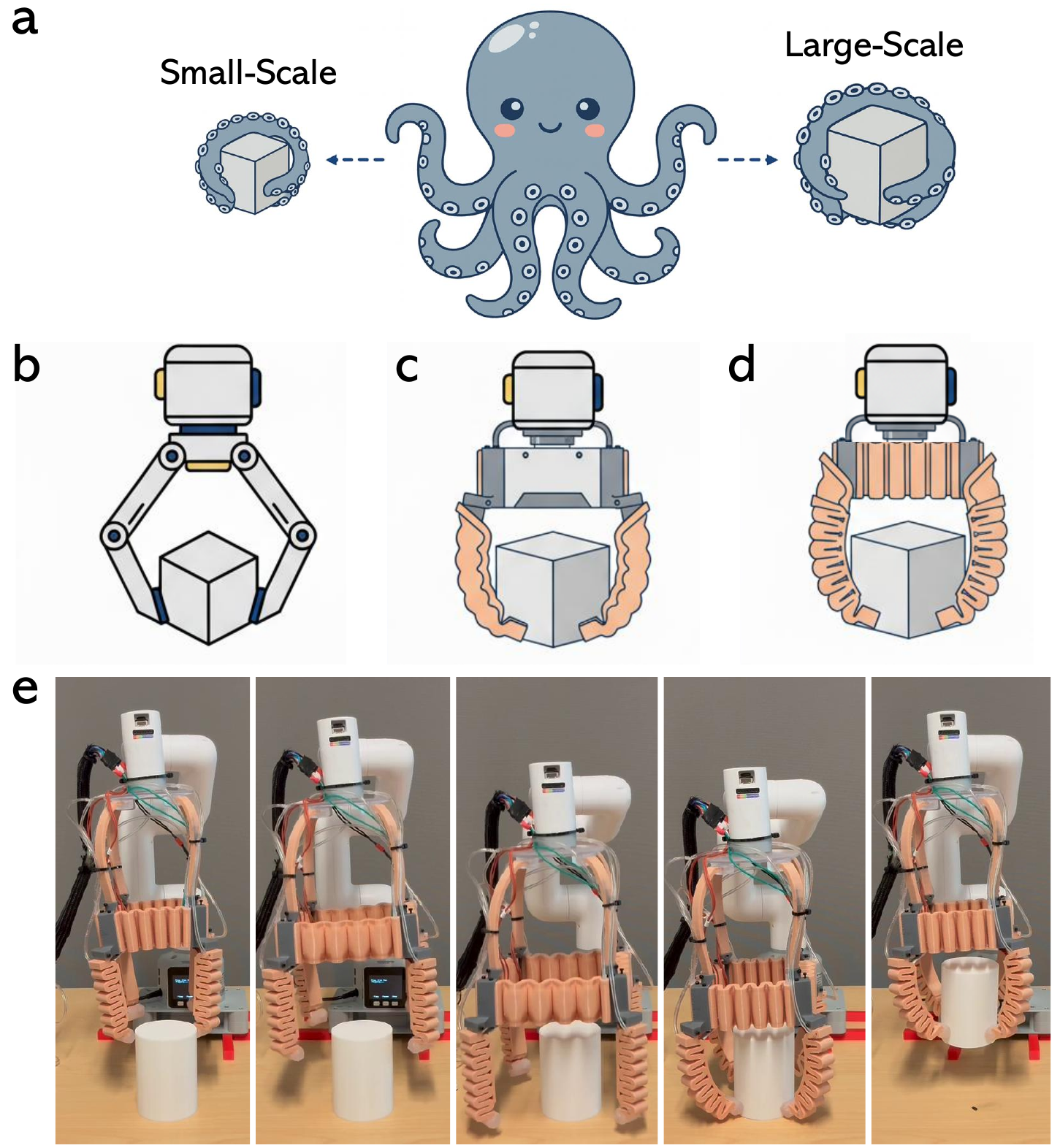}\vspace{-2mm}
\caption{\textcolor{black}{\textbf{Motivation.} 
(\textbf{a}) The octopus uses its flexible tentacles for masterful multi-scale object sensing and manipulation, providing our biological inspiration. 
(\textbf{b}) A rigid gripper fails to conform to the object, resulting in an unstable grasp. 
(\textbf{c}) A conventional soft gripper, limited by its fixed structure, cannot handle objects outside its designed size range. 
(\textbf{d}) Our proposed gripper uses \textbf{\textit{adaptive, whole-body shape morphing}} to reconfigure its entire structure, enabling it to securely envelop objects of varying scales. 
(\textbf{e}) Dynamic grasping sequence: the gripper first adapts its global shape during approach and then performs a final enveloping grasp for a secure lift.}}
\label{fig:teaser}\vspace{-2mm}
\end{figure}

Biological systems, such as the octopus, exhibit a masterful dexterity for cross-scale manipulation, capable of delicately handling a tiny shell one moment and securely engulfing a large, irregular rock the next, as conceptualized in {Fig. \ref{fig:teaser}a}. This remarkable adaptability stems from their ability to perform \textbf{whole-body proprioceptive morphing}, \textit{i.e.}, a capability fundamentally absent in conventional robotics \cite{calisti2011octopus, laschi2016soft, trivedi2008soft, kim2013soft}. While rigid grippers excel in structured industrial automation \cite{reddy2013review}, their fixed kinematics render them brittle when faced with real-world uncertainty ({Fig. \ref{fig:teaser}b}) \cite{teeple2022multi}. This limitation motivated a shift toward soft robotics, which leverages material compliance to passively conform to objects \cite{rus2015design, shintake2018soft}. Yet, despite advancements, the vast majority of soft grippers are still constrained by a \textbf{fixed global morphology}. As illustrated in {Fig. \ref{fig:teaser}c}, this static base structure inherently limits their operational range, causing grasp failure when an object exceeds its designed envelope \cite{11020847, 11020966, 10521930, 10522020}. To overcome this critical bottleneck, we argue that a gripper must be able to reconfigure its entire form factor. Our proposed design ({Fig. \ref{fig:teaser}d}) achieves this robust, cross-scale grasping through \textbf{collaborative, whole-body reconfiguration}, first adapting its global shape during approach and then executing a final enveloping grasp ({Fig. \ref{fig:teaser}e}).

\begin{figure*}[t!]
    \centering    \includegraphics[width=\linewidth]{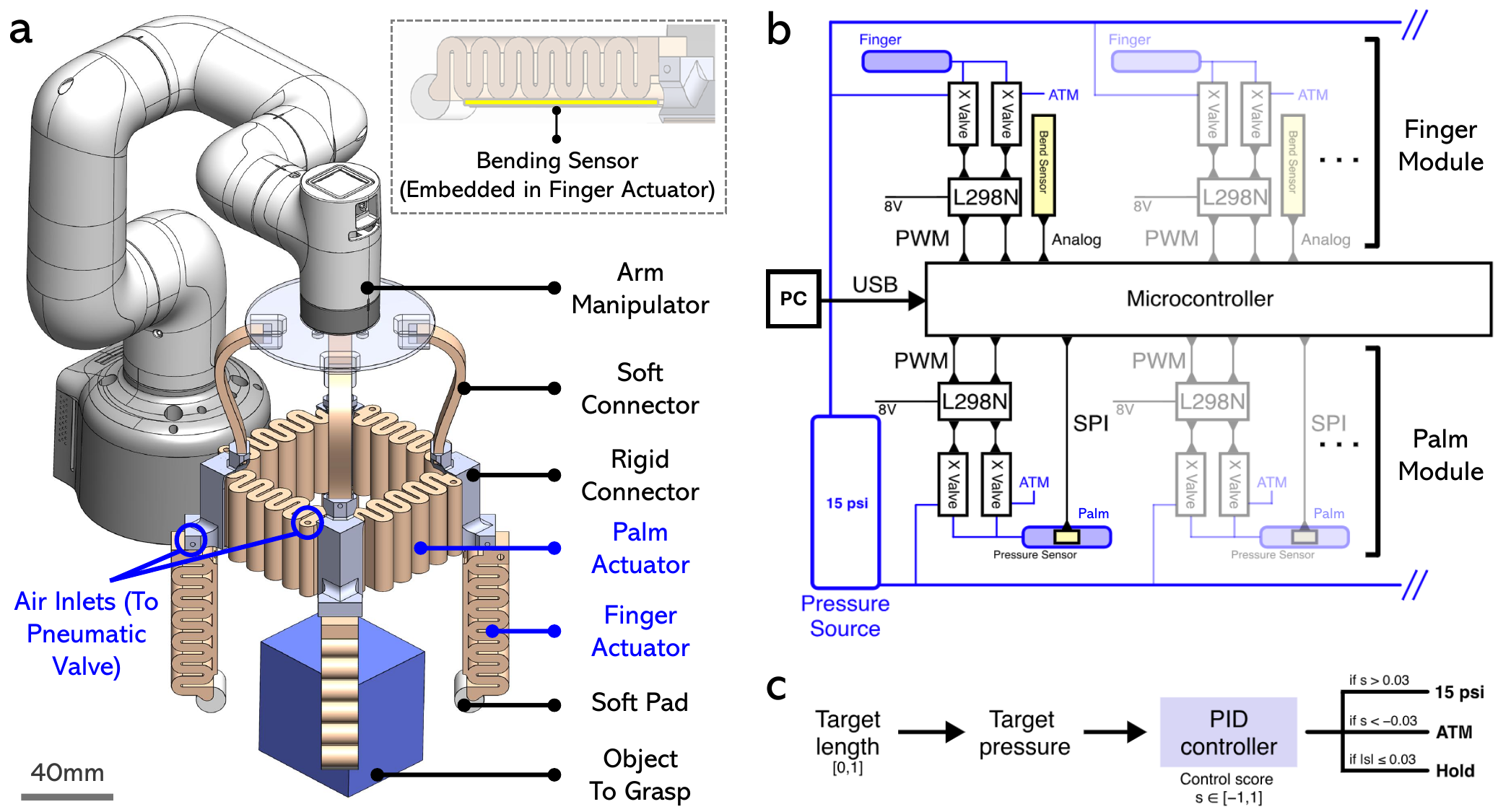}
    \caption{\textbf{System architecture for whole-body proprioceptive  morphing.} 
(\textbf{a}) The modular mechanical design consists of four  \textbf{\textit{morphing palm actuators}} that control the gripper's global shape and four \textbf{\textit{grasping finger modules}} for object envelopment. These are coupled by rigid connectors to form a reconfigurable structure. 
(\textbf{b}) The hierarchical control and sensing schematic. A central microcontroller runs parallel control loops for the finger and morphing modules. \textbf{\textit{Proprioceptive feedback}} is achieved via integrated bend sensors in the fingers and pressure sensors in the morphing actuators. 
(\textbf{c}) Closed-loop pressure control logic for a single morphing actuator. A high-level command for a desired length ($L_{des}$) is translated by a PID controller into a target pressure ($P_{target}$). A low-level logic then modulates pneumatic valves (inflate, hold, deflate) to precisely regulate the actuator's state.}

    \label{fig:system_diagram}
\end{figure*}

The pursuit of adaptable grasping has driven developments across diverse actuation modalities. Tendon-driven mechanisms \cite{gunderman2022tendon} and variable-stiffness systems \cite{zhao2023palm} offer enhanced conformity and stability but remain constrained by a fixed global geometry. While origami-based structures \cite{li2019vacuum} and SMA-actuated designs \cite{lee2019long} introduce reconfigurability, they face trade-offs in force output, control complexity, or response speed. Pneumatic actuation remains a highly practical approach, providing intrinsic compliance, large forces, and simple control \cite{abondance2020dexterous}. However, even within this domain, most designs preserve a fixed morphology \cite{guo2018soft, jain2023multimodal, teeple2020multi}, motivating architectures that can fundamentally reconfigure their shape.

Existing shape-morphing grippers have begun to address this, but often with significant limitations. Many efforts are confined to local shape variation, such as reconfigurable fingertips or adaptive palms, without altering the gripper's overall topology \cite{11020894, 10521918}. Other designs, while innovative, lack practical scalability or feedback. For example, Abondance et al. \cite{abondance2020dexterous} demonstrated localized deformation around a rigid palm, but its mold-based fabrication constrained scalability and it operated in open-loop. Similarly, the vacuum-driven origami of Li et al. \cite{li2019vacuum} lacked sensing and modular reconfiguration. Even recent frameworks embedding actuation and sensing, such as the thermally driven morphing by Sun et al. \cite{sun2023embedded}, are restricted by slow morphing speeds and complex, monolithic fabrication. These studies collectively highlight a critical gap: \textbf{the need for a scalable, reconfigurable, and sensor-integrated design that is also {low-cost and easy-to-fabricate}}.

In contrast, this work introduces a distributed network of modular, self-sensing actuators that enables whole-body morphing and  adaptation. Our architecture is  {low-cost and easy-to-fabricate}, leveraging 3D printing for the rapid assembly of integrated actuator-sensor modules. This modularity allows the structure to generate {multiple, precisely controllable morphing states}—from linear to complex polygonal configurations—without external fixtures. Crucially, we integrate {proprioception with morphing}; embedded self-sensing within each module facilitates closed-loop feedback for intelligent, adaptive reconfiguration. We term this concept {\textbf{proprioceptive morphing}}, \textit{i.e.}, the fusion of distributed actuation and embedded sensing to achieve intelligent adaptation.

This paper makes the following key contributions:

\begin{itemize}
\item We introduce a novel soft gripper architecture based on a distributed network of modular actuators that enables {collaborative, whole-body morphing} for robust multi-scale grasping.
\item We demonstrate the seamless integration of {proprioceptive feedback} via embedded self-sensing of the grasping finger actuators.
\item We present a fully modular and accessible design, leveraging 3D printing for the {rapid and low-cost fabrication} of integrated soft actuator–sensor modules.
\item We validate the superior performance of our gripper, showcasing its ability to robustly and securely grasp objects across a wide range of sizes and shapes.
\item We demonstrate that our gripper's architecture unlocks novel manipulation strategies, including simultaneous multi-object grasping and internal hook grasping, which are infeasible for conventional designs.
\end{itemize}

\section{System Design for Proprioceptive Morphing}

\textcolor{black}{Our gripper's ability to achieve robust, cross-scale grasping is rooted in a design philosophy of \textbf{distributed actuation and integrated proprioception for whole-body morphing}. Instead of a monolithic structure, we designed a modular system where distinct functional units collaborate to achieve complex behaviors. This section details the hardware architecture and control framework that enable our gripper's whole-body morphing capabilities, as illustrated in Fig. \ref{fig:system_diagram}.}

\subsection{Modular Mechanical Architecture}

\begin{figure}[t]
    \centering
    \includegraphics[width=\linewidth]{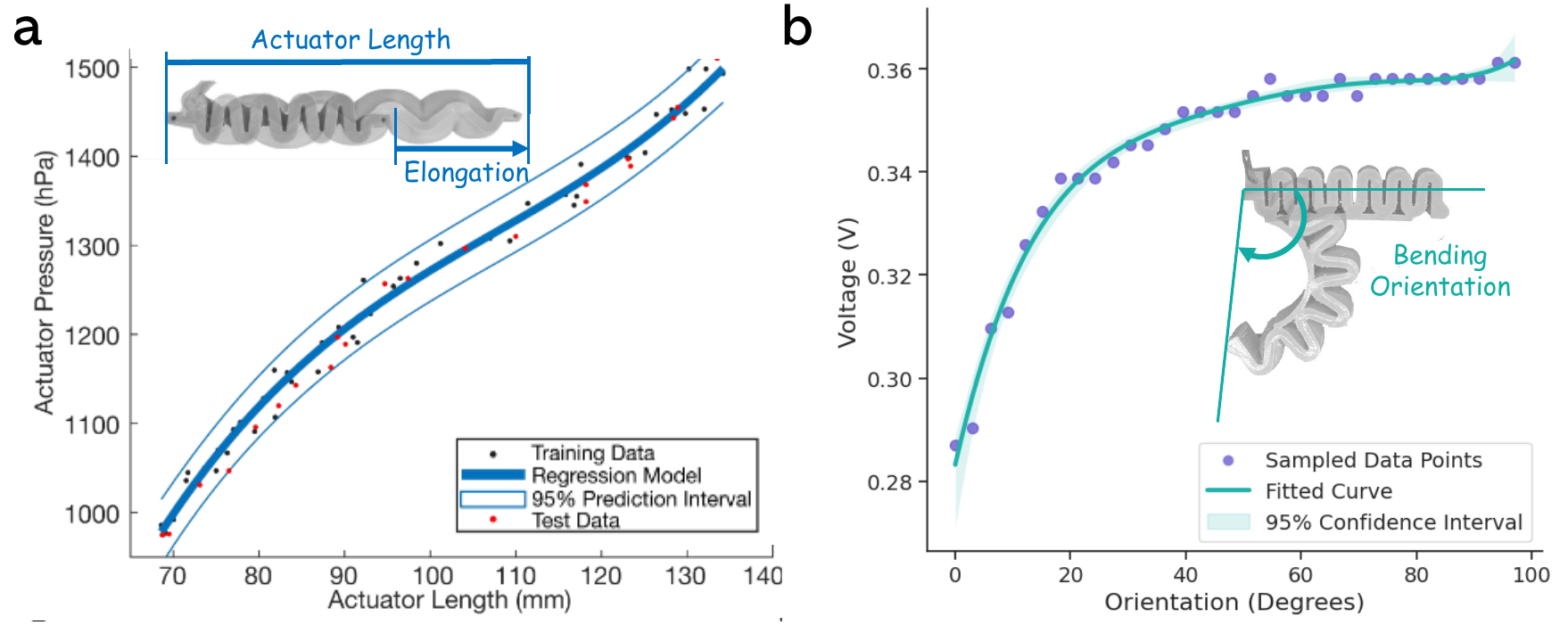}\vspace{-2mm}
 \caption{\textbf{Single actuator characterization.} The plots show a predictable relationship between applied pressure and the resulting (\textbf{a}) length of the palm actuator and (\textbf{b}) bending angle of the finger actuator. Illustrations show the corresponding shape morphing transitions: the palm extends from ~68 mm to ~135 mm to reconfigure and morph the gripper's framework shape, while the finger bends inward to grasp objects. }
    \label{fig:actuator_characterization}
\end{figure}

\textcolor{black}{The gripper is a modular assembly of two distinct types of soft pneumatic actuators, as shown in {Fig. \ref{fig:system_diagram}a}.
\noindent\textbf{1) Morphing Palm  Actuators:} Four linear PneuFlex-style actuators form the reconfigurable framework of the gripper. These actuators were designed following a kirigami-inspired PneuNet architecture\cite{guo2023kirigami}. Their primary function is to control the gripper's \textbf{\textit{global shape and size}}. By precisely controlling their extension, the gripper can dramatically alter its workspace, transitioning from a compact configuration for small objects to a wide-open configuration for large targets, as shown in Fig.~\ref{fig:actuator_characterization}a. \textbf{2) Grasping Finger Actuators:} Four bending actuators perform the \textbf{\textit{local object envelopment and grasping}}. Positioned at the corners of the morphing framework, they are responsible for making compliant contact and securely conforming to the object's local geometry, as shown in Fig.~\ref{fig:actuator_characterization}b. }

\begin{figure}[t]
    \centering
    \includegraphics[width=\linewidth]{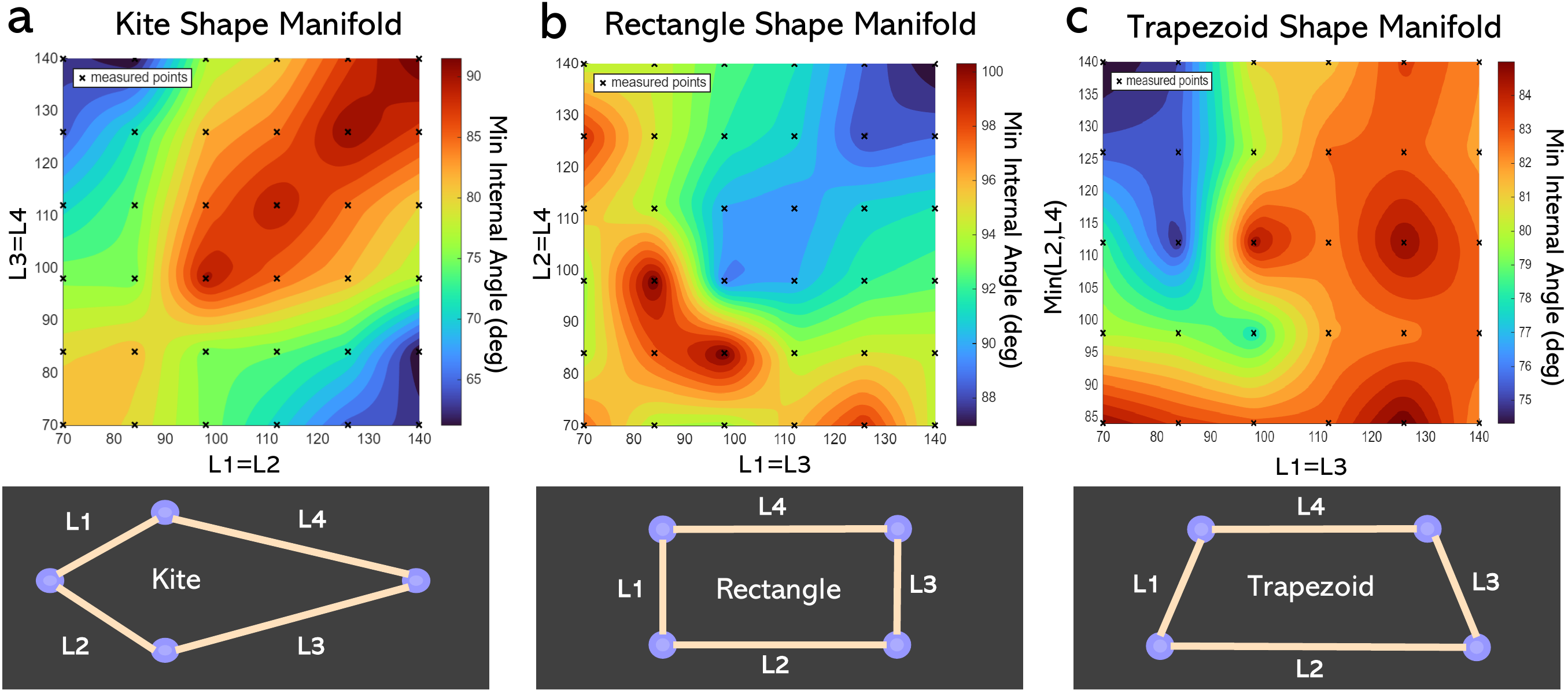}\vspace{-2mm}
\caption{\textbf{Shape manifold characterization of the framework formed by the four palm actuators.} Interpolated heatmaps illustrating the palm module's ability to achieve various target shapes. For each shape – (\textbf{a}) kite, (\textbf{b}) rectangle, and (\textbf{c}) trapezoid – the heatmaps show the minimum internal angle (color bar) as a function of adjustable link lengths (x and y axes in millimeter scale, as defined in the accompanying diagrams). The `x' marks indicate the measured points used for interpolation.}
    \label{fig:shape_manifold}
\end{figure}

\textcolor{black}{All modules are fabricated using multi-material 3D printing with flexible TPU filament, enabling rapid prototyping and customization. This modularity is a key design feature that allows easy assembly, repair, and scalability. As shown by the characterization data in {Fig. \ref{fig:actuator_characterization}}, both actuator types exhibit a predictable and repeatable relationship between input pressure and displacement, which is essential for precise control. The ability to control the palm actuators allows the system to generate a wide manifold of shapes, including kites, rectangles, and trapezoids, enabling it to pre-shape itself for a diverse range of objects (see {Fig. \ref{fig:shape_manifold}}).}

\subsection{Fabrication Process}

\begin{figure}[t]
    \centering
    \includegraphics[width=\linewidth]{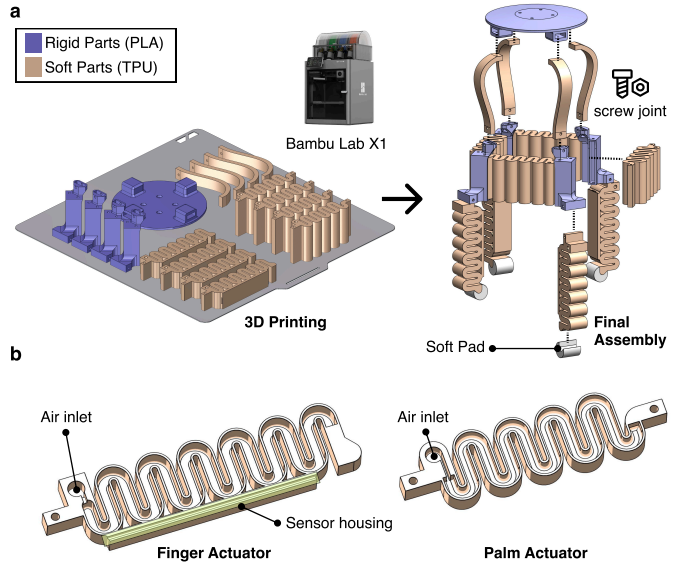}\vspace{-2mm}
\caption{\textbf{Fabrication process.} (\textbf{a}) All parts are fabricated using two materials, TPU and PLA, and printed using a Bambu Lab X1 printer. The entire set can be printed simultaneously on a single 256 mm × 256 mm build plate over approximately 36 hours of printing. After 3D printing, all components are assembled using screw joints, except for the friction pads at the fingertip, which are bonded with silicone epoxy (Silpoxy). (\textbf{b}) Cross-sectional views of the finger actuator and palm actuator are shown. The yellow-shaded regions indicate the sensor housings, and pneumatic pressure is applied to both actuators through the air inlet ports. The holes at the distal ends correspond to M2 screw locations used for fastening.}
    \label{fig:fabrication}
\end{figure}

Our adaptive gripper is a low-cost and easily fabricated soft robotic system designed for accessibility, reproducibility, and educational use. With a total component cost of approximately \$105, the entire system can be rapidly built using only a desktop 3D printer and widely available commercial off-the-shelf (COTS) components. As summarized in Table~\ref{tab:bom}, all structural parts, including actuators, connectors, and housings, are 3D-printed using TPU and PLA on a Bambu Lab X1 printer. The complete set fits within a single 256 mm × 256 mm build plate and can be printed simultaneously in roughly 36 hours.

After printing, all components are assembled through screw joints without specialized tools, enabling quick reconfiguration or repair. Only the fingertip friction pads require adhesive bonding, achieved using silicone epoxy (Silpoxy). Fig.~\ref{fig:fabrication} illustrates the fabrication process and assembly sequence, along with cross-sectional views of the finger and palm actuators. The yellow-shaded regions indicate the sensor housings, and pneumatic pressure is supplied through designated air inlets. Each distal end includes M2 screw holes for fastening.

This streamlined fabrication pipeline highlights the gripper’s practical novelty: it can be manufactured, assembled, and operated with minimal resources, making it an ideal platform for both research and educational applications in soft robotics.

\subsection{Distributed Sensing and Electronics}

\textcolor{black}{To achieve intelligent, adaptive object grasping behavior, the mechanical architecture is augmented with a distributed network of sensors, providing rich {proprioceptive contact feedback}, as diagrammed in {Fig. \ref{fig:system_diagram}b}.  {Grasping fingers} have integrated flexible bending sensors laminated along their neutral axis to measure their curvature. This serves as a proxy for both their own configuration and for detecting contact with an object when conducting grasping.}

\textcolor{black}{This sensor suite allows the gripper to feel its own contact state and its interaction with the environment. The electronic system is designed to be both low-cost and robust. A central Raspberry Pi Pico microcontroller manages all control loops and sensor acquisition, and a custom-designed PCB integrates L298N drivers to operate eight solenoid valves, providing independent pneumatic control for each module. }

\subsection{Hierarchical Control for Adaptive Grasping}

\textcolor{black}{The gripper's operation is governed by a hierarchical control strategy that separates low-level actuator regulation from high-level grasping logic.}

\textbf{Low-Level Actuator Control.} At the lowest level, each of the eight modules is managed by an independent closed-loop controller. As shown in {Fig. \ref{fig:system_diagram}c}, a high-level command (e.g., a desired length $L_{des}$) is fed to a PID controller. The controller's output is a target pressure $P_{target}$, which is achieved by modulating the corresponding solenoid valve to inflate, hold, or deflate the actuator. This allows for precise and stable control over the state of each actuator module.

\begin{table}[t!]
\centering
\caption{Bill of Materials for our Low-Cost, 3D-Printed Soft Gripper System with commercial off-the-shelf (COTS) components.}
\label{tab:bom}
\resizebox{\linewidth}{!}{%
\begin{tabular}{@{}llccc@{}}
\toprule
\textbf{Component} & \textbf{Category} & \textbf{Qty.} & \textbf{Fabrication/Source} & \textbf{Est. Unit Cost} \\ \midrule
\multicolumn{5}{c}{\textit{\textbf{Mechanical Components (Primarily 3D-Printed)}}} \\
All Actuators \& Links & Mechanical & - & 3D-Printed & \textless \$10 (Filament) \\
\midrule
\multicolumn{5}{c}{\textit{\textbf{Control \& Electronics (Low-Cost COTS)}}} \\
RP Pico & Microcontroller & 1 & COTS (Low-Cost) & \$5.00 \\
L298N Driver & Motor Driver & 8 & COTS (Low-Cost) & \$1.50 \\
Bending Sensor & Sensing & 4 & COTS / Custom & \$3.00 \\
SPI Pressure Sensor & Sensing & 4 & COTS & \$4.00 \\
\midrule
\multicolumn{5}{c}{\textit{\textbf{Pneumatic System (Low-Cost COTS)}}} \\ 
Solenoid Valve & Pneumatics & 8 & COTS (Low-Cost) & \$8.00 \\
Pressure Source & Pneumatics & 1 & COTS (15 psi pump) & \$15.00 \\
Pneumatic Tubing & Pneumatics & 1 lot & COTS & \$5.00 \\
\midrule
\multicolumn{4}{r}{\textbf{Total Estimated System Cost}} & \textbf{\$105} \\ \bottomrule
\end{tabular}} \vspace{-2mm}
\end{table}

\textbf{High-Level Adaptive Grasping Strategy.} Building upon the low-level control, we implement a three-phase adaptive grasping policy that leverages the gripper's unique whole-body morphing capability. The strategy is designed to be robust and sensor-driven, rather than relying on a precise model of the object.
\noindent\textbf{1) \textit{Phase 0: Grasping Initialization.}} Before each grasping attempt, the gripper is set to a default, open configuration. The palm actuators are fully extended to their maximum length, and the finger actuators are unactuated (i.e., held straight). The robotic arm then lowers the gripper toward the target object to prepare for the adaptive enveloping process.
\noindent\textbf{2) \textit{Phase 1: Global Reconfiguration (Approach)}.} As the gripper approaches a target object, the four {palm morphing actuators} are actuated to pre-shape the gripper's framework. The goal is to match the overall scale of the object, ensuring all four grasping fingers are in a position to make effective contact. This phase primarily uses an open-loop strategy based on a rough estimate of the object's size, leveraging the characterized performance shown in \textbf{Fig. \ref{fig:actuator_characterization}a}. \noindent\textbf{3) \textit{Phase 2: Envelopment and Secure Grasp (Contact)}.} Once the gripper is pre-shaped and positioned, the four {grasping finger modules} are actuated. Their inflation is governed by a closed-loop, contact-driven protocol. Each finger inflates until its embedded bend sensor reading surpasses a pre-defined threshold, indicating firm contact. Upon detection of contact, the action of that finger is stopped to prevent the application of excessive force and to ensure a stable, form-fitting grip. The grasp is considered successful and secure when a majority (e.g., 3 out of 4) of the fingers have reported contact. This distributed, event-driven closure mimics biological reflexes and results in a highly adaptive and secure enveloping grasp.

\subsection{Contact State Detection from Proprioceptive Feedback}
To achieve a robust, contact-driven grasping strategy, we implemented an automated method to detect the precise moment of transition from a non-contact to a contact state using the proprioceptive feedback from the finger bending sensors. This process is critical for the closed-loop feedback described in Phase 2 of our grasping policy. The detection algorithm proceeds as follows: First, the time-series data from each of the four finger sensors is normalized by subtracting its initial value at time step zero, making the analysis relative to the starting configuration. To reduce signal noise and prevent spurious detections, a median filter with a kernel size of 5 is applied to each normalized signal. 

The core of the detection logic lies in analyzing the rate of change of the filtered signal. We compute the first-order difference between consecutive time steps, and a contact event is registered for an individual finger when the absolute value of this difference exceeds a predefined threshold (\textit{TRANSITION\_THRESHOLD = 5.0}). This sharp change indicates a rapid increase in curvature due to physical contact with the object. The definitive transit point for the entire gripper is then determined as the earliest time step at which any of the four fingers first registers such a contact event. This event-driven approach ensures the gripper responds at the initial moment of firm contact, enabling an adaptive grasp, as visualized by the dashed line in the Fig.~\ref{fig:sensor_readings_grid} of the experiment section.

\begin{figure}[t]
    \centering
    \includegraphics[width=\linewidth]{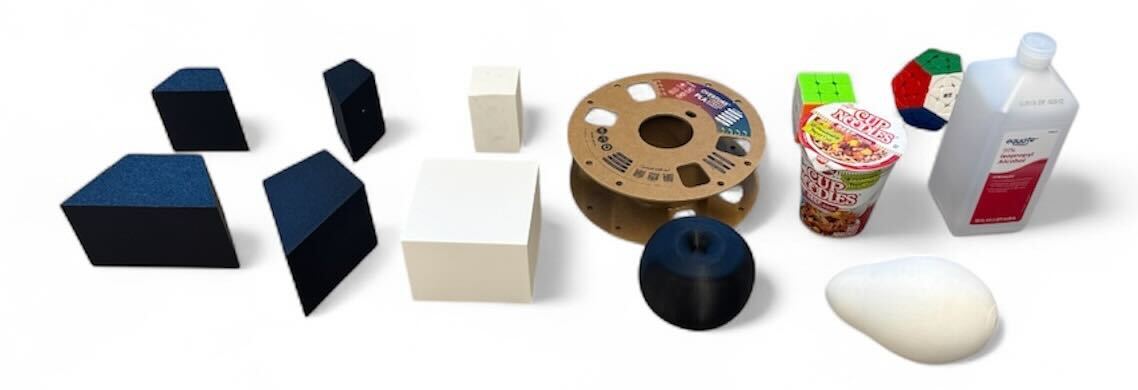}\vspace{-2mm}
\caption{\textbf{Tested objects for grasping capability analysis}. Our grasping experiments utilized a diverse object set, including 3D-printed items (of both standard and irregular shapes) and various real-world objects. }
    \label{fig:tested_objects}\vspace{-2mm}
\end{figure}

\begin{table}[t!]
\centering
\caption{Geometric Parameters of Test Objects.}
\label{tab:object_params}
\renewcommand{\arraystretch}{1.0} 
\begin{tabular*}{0.35\textwidth}{ @{\extracolsep{\fill}}lc@{} }
\toprule
\textbf{Object Name} & \textbf{Dimensions (mm)} \\ \midrule
\multicolumn{2}{c}{\textit{Standardized Geometric Objects}} \\ \midrule
Kite (Small) & $80 \times 50 \times 80$ \\
Kite (Large) & $140 \times 90 \times 80$ \\ \addlinespace[0.5em]
Rectangle (Small) & $20 \times 20 \times 80$ \\
Rectangle (Large) & $110 \times 110 \times 80$ \\ \addlinespace[0.5em]
Trapezoid (Small) & $80/40 \times 40 \times 80$ \\
Trapezoid (Large) & $140/70 \times 70 \times 80$\\ \midrule
\multicolumn{2}{c}{\textit{Complex-Shape \& Real Objects}} \\ \midrule
3D-Printed Pear &  $95 \times 95 \times 160$\\
6-Face Cube & $56 \times 56 \times 56$\\
10-Face Cube & $90 \times 90 \times 90$\\
Cup Noodles &  $90 \times 90 \times 108$\\ 
Bottle &  $80 \times 80 \times 250$\\ 
3D-Printed Apple &  $105 \times 105 \times 93$\\ 
Delivery Box & $87 \times 87 \times 118$\\
3D Printing Tray & $200 \times 200 \times 64$\\

\bottomrule
\end{tabular*} \vspace{-2mm}
\end{table}

\section{Experimental Validation}

We conducted a series of experiments to rigorously evaluate the performance of our proprioceptive morphing gripper. The evaluation was designed to answer three primary questions: 1) Is whole-body morphing necessary for robust cross-scale grasping? 2) How versatile and effective is our adaptive gripper on a diverse range of objects? 3) Can our proprioceptive sensing distinguish between gripper-object contact states?

\subsection{Experimental Setup}

To provide a quantitative basis for our grasping experiments, we defined a set of test objects with specific geometries, as illustrated in Fig.~\ref{fig:tested_objects} and detailed in Table~\ref{tab:object_params}. 
For each test object, three characteristic dimensions were defined: maximum width, maximum depth, and height, measured along the principal orthogonal axes of the object.
For trapezoidal objects, four dimensions were specified: bottom width, top width, depth, and height.
The bottom and top widths correspond to the longer and shorter parallel edges of the trapezoidal cross section, respectively.
The standardized objects were 3D-printed with controlled dimensions to systematically evaluate the gripper's performance against variations in shape and scale. The real-world objects were selected to represent a diverse range of common items with varying shapes and sizes, allowing for a practical assessment of the gripper's versatility. Notice that the largest object, \textit{i.e.}, a 3D printing tray, and the smallest object, \textit{i.e.}, a small rectangle-shaped object, have a relative size ratio of 10.

All grasping trials were performed with the gripper mounted as an end effector on a 6-DoF robotic arm to ensure repeatable positioning and execution. We used two sets of test objects: 1) a standardized set of 3D-printed geometric primitives (kites, rectangles, trapezoids) of two distinct sizes (Small/Large) to systematically probe the effects of scale and shape, and 2) a set of common real-world objects (\textit{e.g.}, a pear, cube, a cup of noodles) to evaluate practical utility. A grasp is  a success if the object could be safely lifted from the work surface and held for 5 seconds without any observable slippage.

\begin{table}[t!]
\centering
\caption{Manipulation Success Rate for Different Object Shapes and Gripper Configurations }
\label{tab:manipulation_success}
\begin{tabular}{@{}l|cc|cc|cc@{}}
\toprule
\multirow{3}{*}{\textbf{Palm Configuration}}  & \multicolumn{6}{c}{\textbf{Tested Standard Objects}}\\ 
\cmidrule{2-7}
& \multicolumn{2}{c|}{Kite} & \multicolumn{2}{c|}{Rectangle} & \multicolumn{2}{c}{Trapezoid} \\
\cmidrule{2-7}
 & S & L & S & L & S & L \\
\midrule
Gripper-Rec-S & \cmark & \xmark & \cmark & \xmark & \cmark & \xmark \\
Gripper-Rec-L & \xmark & \cmark & \xmark & \cmark & \xmark & \cmark \\
Gripper-Trapez-S & \xmark & \xmark & \xmark & \xmark & \cmark & \xmark \\
Gripper-Trapez-L & \xmark & \xmark & \xmark & \xmark & \xmark & \cmark \\
Gripper-Kite-S & \cmark & \xmark & \cmark & \xmark & \xmark & \xmark \\
Gripper-Kite-L & \xmark & \cmark & \xmark & \cmark & \xmark & \xmark \\
\bottomrule
\end{tabular}\vspace{-2mm}
\end{table}

\begin{table}[t!]
\centering
\caption{Grasping Success Rate on Real-World Objects with a Dynamically Morphing Palm.}
\label{tab:real_world_grasping}
\begin{tabular}{@{}l|ccc@{}}
\toprule
\multirow{2}{*}{\textbf{Palm Configuration}}  & \multicolumn{3}{c}{\textbf{Real-World Objects}} \\
\cmidrule(l){2-4}
 & Pear & Cube & Cup Noodle \\ \midrule
Rectangle & \cmark & \cmark & \xmark \\
Trapezoid & \cmark & \cmark & \xmark \\
Kite & \xmark & \cmark & \cmark \\ \bottomrule
\end{tabular} \vspace{-4mm}
\end{table}

\subsection{Validating the Need for Whole-Body Morphing}

\textcolor{black}{To establish a baseline and validate the core hypothesis of our work, we first evaluated the performance of the gripper with its palm locked into fixed configurations, effectively mimicking conventional non-adaptive soft grippers. We tested six fixed palm shapes against our standardized object set. The results, summarized in {Table \ref{tab:manipulation_success}}, lead to two conclusions.}

\textcolor{black}{\textbf{First, no single fixed configuration could successfully grasp all objects}, unequivocally demonstrating the limitations of a fixed morphology. For instance, the small rectangular palm configuration (\texttt{Gripper-Rec-S}) successfully grasped all small objects but failed on all large ones due to its limited scale. Conversely, the large configuration (\texttt{Gripper-Rec-L}) secured the large objects but failed on the small ones, often by being unable to properly constrain them. This  validates our motivation: a versatile gripper requires not only compliant fingers but also an adaptive palm to reconfigure its {scale}.}

\textcolor{black}{\textbf{Second, shape matching is critical but limits versatility in fixed designs.} The highly specialized trapezoidal configurations (\texttt{Gripper-Trapez-S/L}) succeeded only with trapezoidal objects, failing on all others. This highlights that while a specific palm shape offers advantages for similar object geometries, it becomes a significant constraint when faced with diversity. Collectively, these baseline experiments confirm that a truly robust and versatile gripper must be capable of dynamically reconfiguring both its {scale and shape}.}

\begin{figure}[t]
    \centering
    \includegraphics[width=\linewidth]{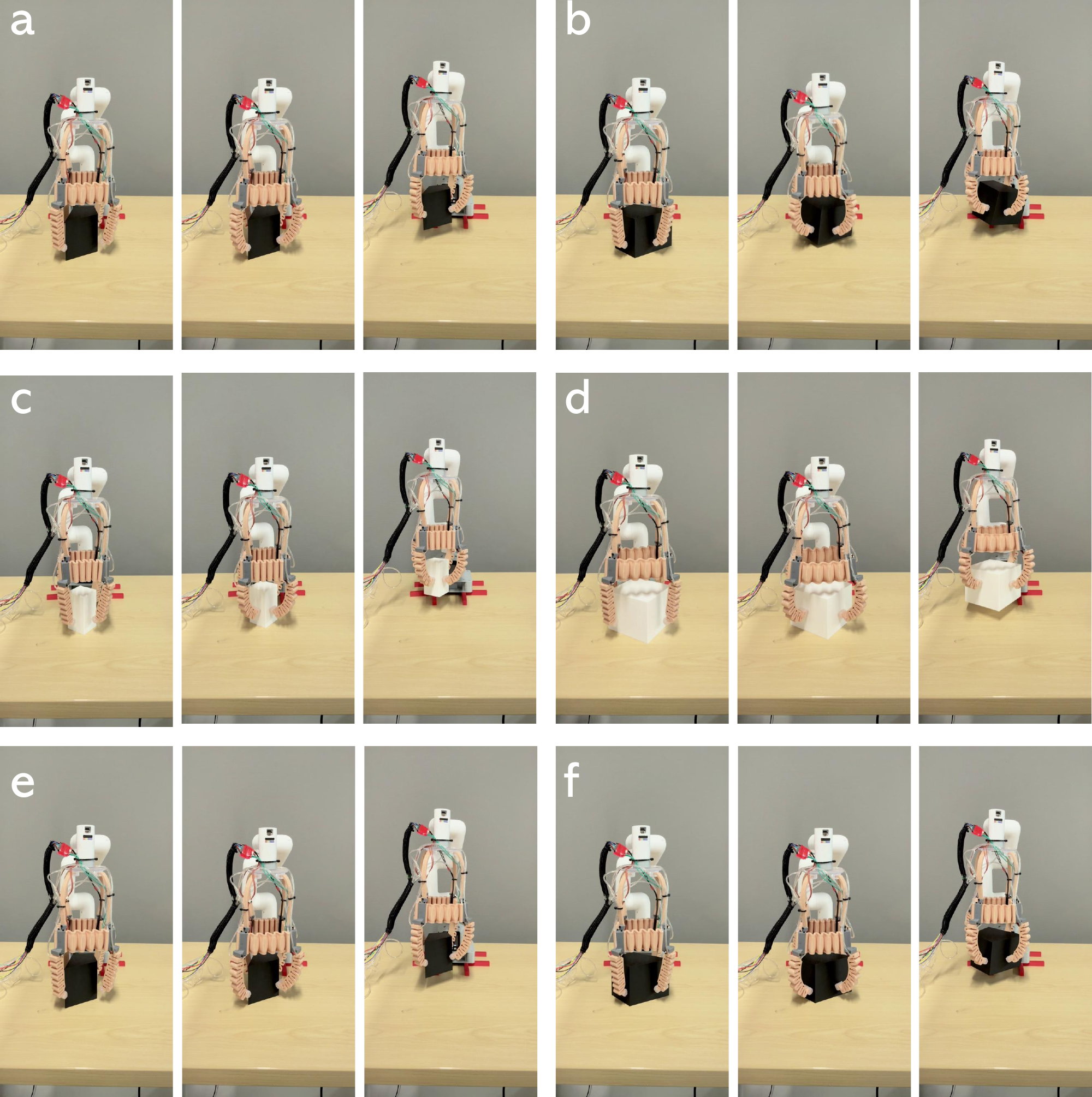}\vspace{-2mm}
\caption{\textbf{Real-world adaptive grasping of various standard objects.} The gripper's grasping sequence is demonstrated on three objects with different cross-sections. For each shape, we illustrate two sizes of objects for grasping demonstration. For each setting, the columns from left to right show the gripper approaching, conforming its shape to the object to establish a grasp, and securely lifting it.}
    \label{fig:grasping_sequence}\vspace{-2mm}
\end{figure}


\begin{figure}[t]
    \centering
    \includegraphics[width=\linewidth]{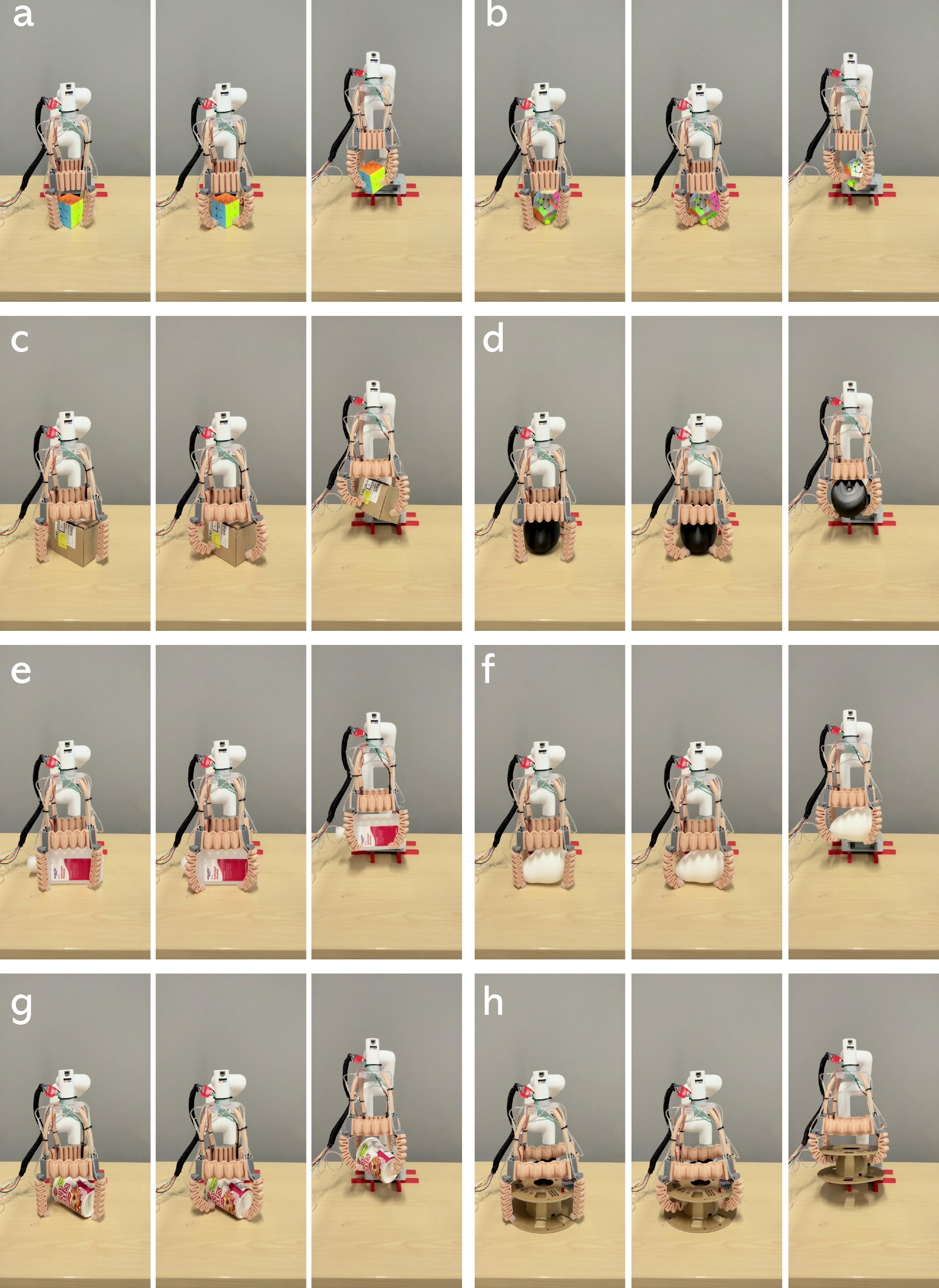}\vspace{-2mm}
\caption{\textbf{Real-world adaptive grasping of various irregular real-world objects.}  For each setting, we show the gripper approaching, conforming its shape to the object to establish a grasp, and securely lifting it.}
    \label{fig:grasping_sequence_complex}
\end{figure}\vspace{-2mm}

\begin{figure*}[htbp] 
    \centering

    \subfloat[Trapezoid, Small]{\includegraphics[width=0.32\textwidth]{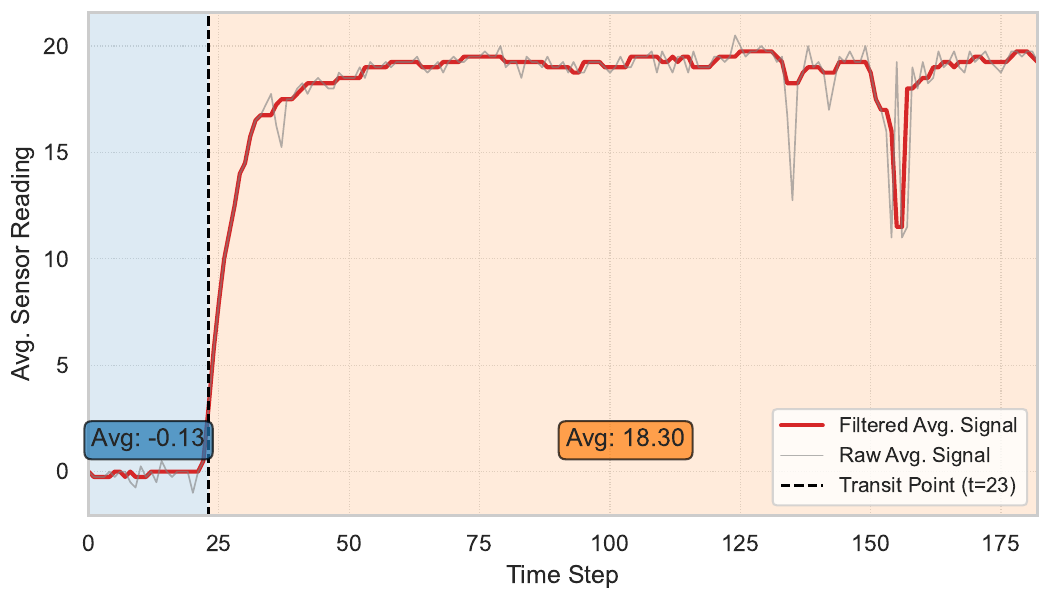}\label{fig:trap_s1}}
    \hfill
    \subfloat[Trapezoid, Large]{\includegraphics[width=0.32\textwidth]{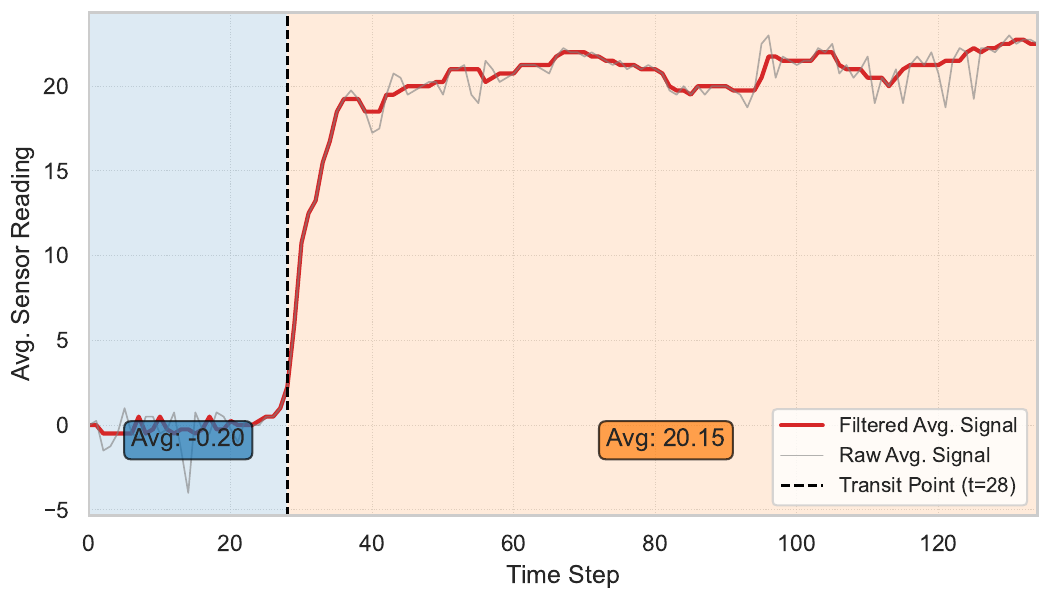}\label{fig:trap_l3}}
    \hfill
    \subfloat[Rectangle, Small]{\includegraphics[width=0.32\textwidth]{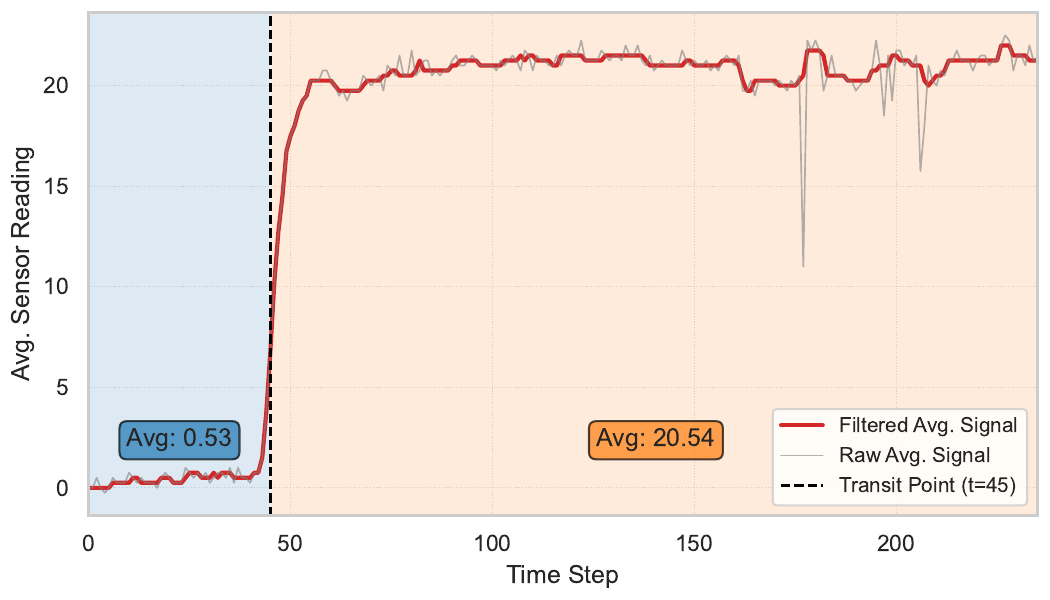}\label{fig:sq_s2}}
    \hfill
    \subfloat[Rectangle, Large]{\includegraphics[width=0.32\textwidth]{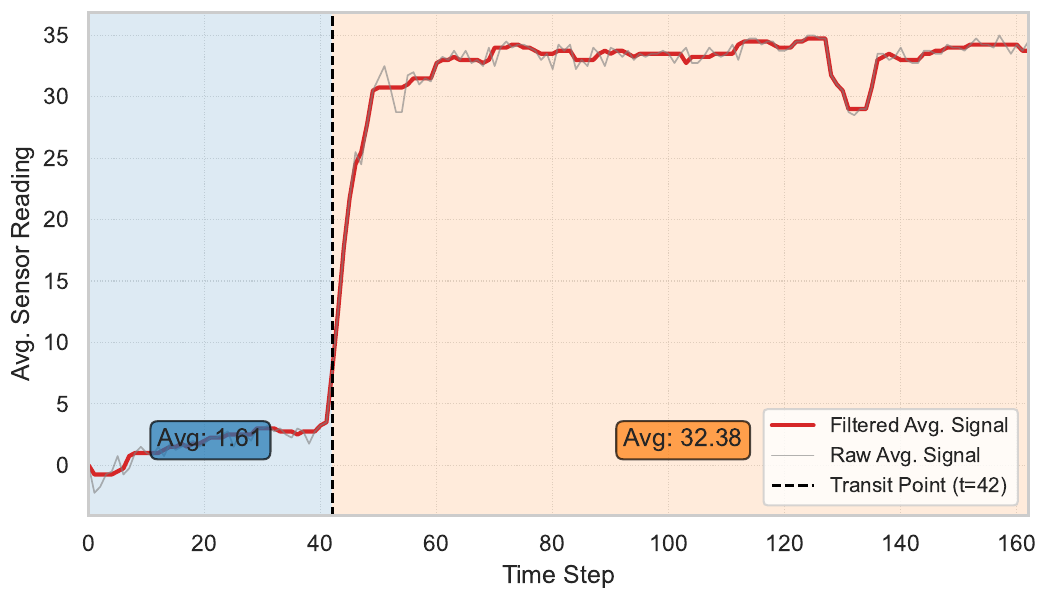}\label{fig:sq_l2}}
\hfill
    \subfloat[Kite, Small]{\includegraphics[width=0.32\textwidth]{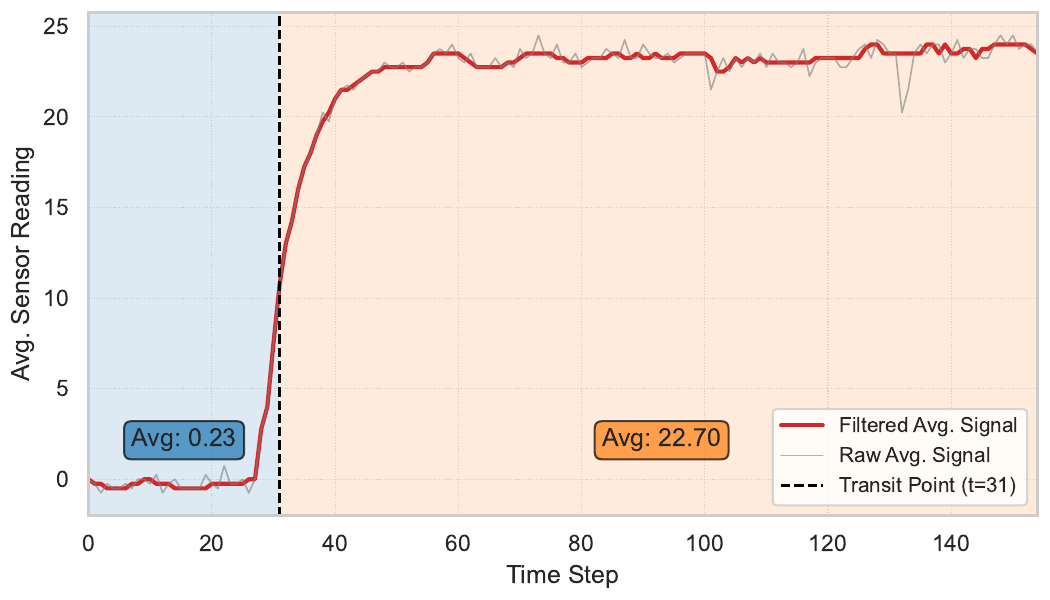}\label{fig:kite_s1}}
    \hfill
    \subfloat[Kite, Large]{\includegraphics[width=0.32\textwidth]{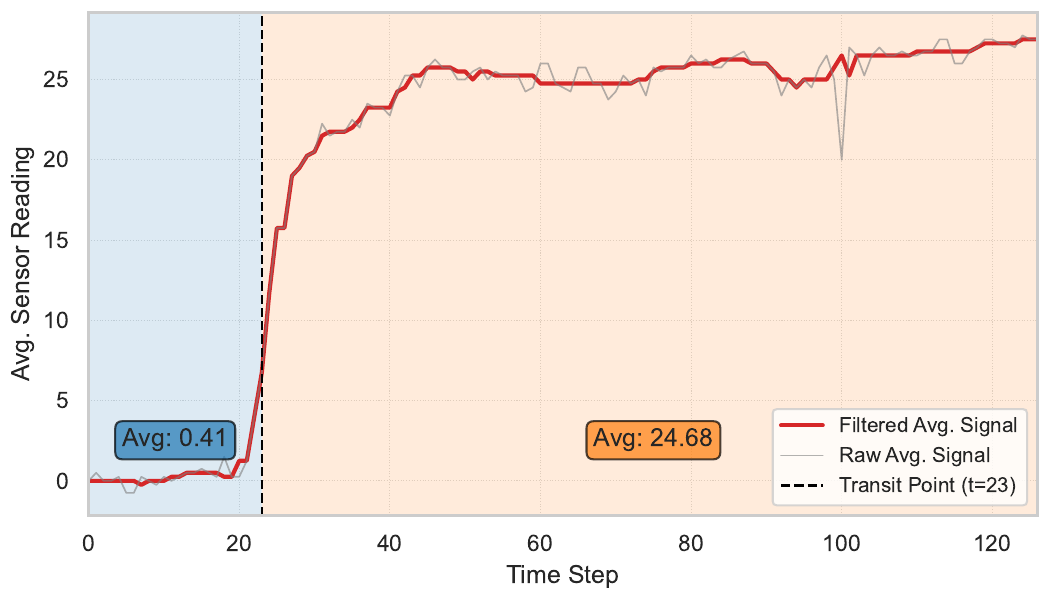}\label{fig:kite_l3}}
    \caption{\textbf{Bending sensor readings for various shape-size configurations during grasping. }Each subplot illustrates the average sensor signal (filtered and raw) relative to its initial value (time step zero) across four finger sensors. A clear transit point, marked by a dashed line, indicates the moment of significant change in the sensor reading. The shaded regions highlight the average signal before and after this transition. }

    \label{fig:sensor_readings_grid}\vspace{-3mm}

\end{figure*}

\subsection{Performance of the Adaptive Morphing}

\textcolor{black}{Having established the limitations of fixed-morphology designs, we then evaluated the full capabilities of our gripper with its dynamic, whole-body morphing enabled.}

\textcolor{black}{\textbf{Versatility on Diverse Geometries.} As shown in the grasping sequences in {Fig. \ref{fig:grasping_sequence}}, the gripper successfully manipulates a diverse set of standard geometries (rectangular, trapezoidal, and kite) and complex real-world objects. In each case, the gripper executes its two-phase adaptive strategy: first, the palm morphs to match the object's overall scale (Phase 1), and then the fingers actuate to achieve a secure, enveloping grasp (Phase 2). This demonstrates the practical effectiveness of our control strategy and the gripper's ability to adapt to different shapes without requiring a tool change.}

\textcolor{black}{\textbf{Real-World Grasping Utility.} To assess the gripper's practical utility, we tested its ability to grasp a set of challenging real-world objects, allowing it to dynamically select the most appropriate palm configuration for each. The results in \textbf{Table \ref{tab:real_world_grasping}} highlight its adaptability. The gripper successfully grasped the irregular pear and the  cube by morphing into rectangular or trapezoidal shapes. Interestingly, only the kite configuration succeeded on the Cup Noodle package. This suggests that the kite shape, with its wider base, provides a more stable caging grasp for cylindrical objects, a task where the other configurations failed.} 
 In Fig.~\ref{fig:grasping_sequence_complex}, we demonstrate the gripper's ability to dynamically reconfigure both its scale and shape to grasp a wide range of objects. These objects vary significantly in geometry, scale, and rigidity.

\subsection{Proprioceptive Contact Sensing of Grasping}
The {ability of our integrated sensing to reflect the state transition from non-contact to contact} is demonstrated in Fig.~\ref{fig:sensor_readings_grid}, which presents the average bending sensor readings during grasping trials across {six} distinct shape-size configurations. These plots clearly validate our contact detection methodology and highlight the quality of the proprioceptive feedback. In each trial, the sensor signal exhibits a distinct two-phase pattern: an initial period of near-zero readings during the gripper's approach, followed by a sharp, unambiguous rise upon making contact with the object, and finally settling into a new, stable, high-value state as the grasp is secured.

Our automated transit point detection algorithm, marked by the vertical dashed line in each subplot, consistently and accurately identifies the moment this state transition occurs.

\begin{figure*}[t]
    \centering
    \includegraphics[width=\linewidth]{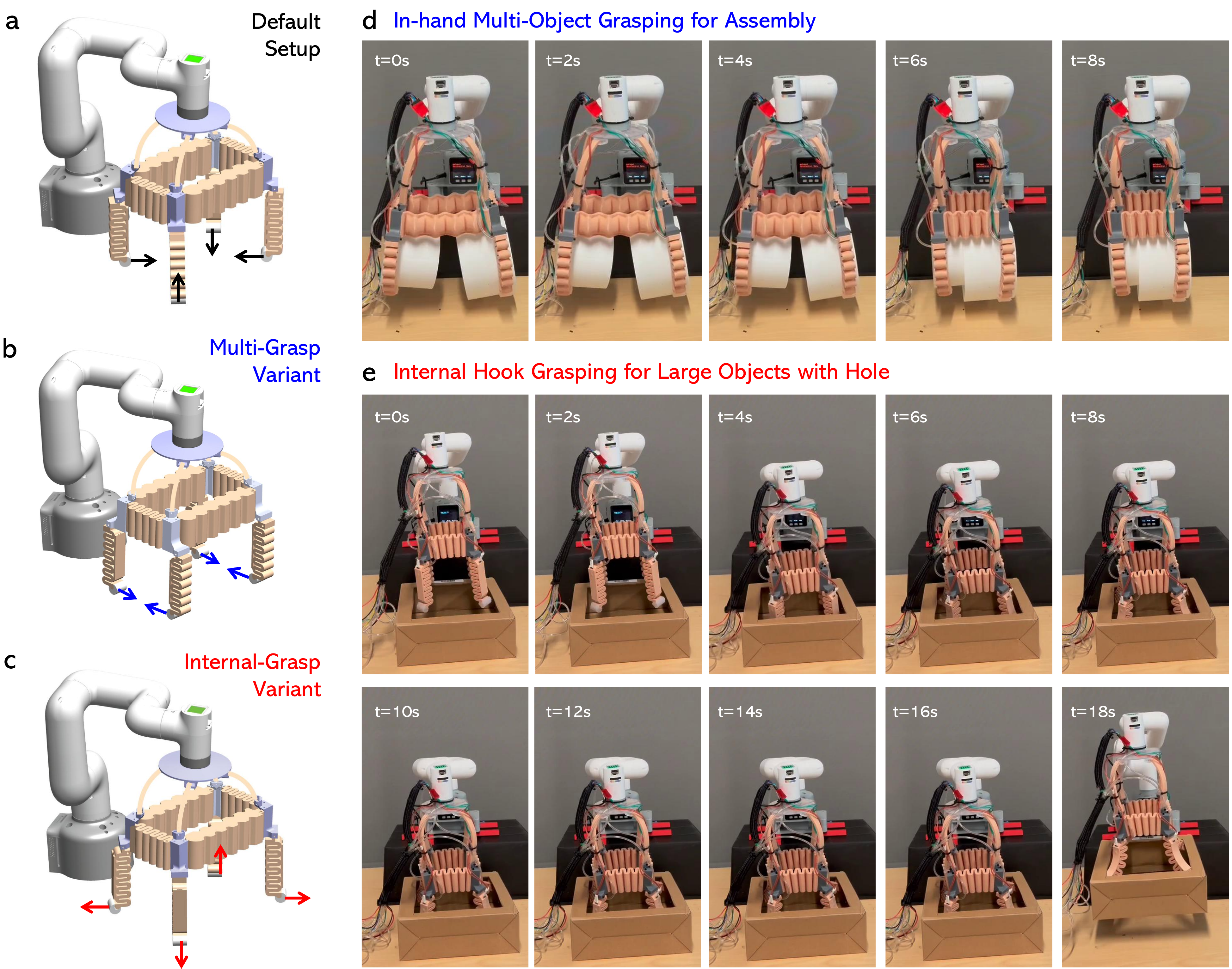}\vspace{-2mm}
\caption{\textbf{ Real-world adaptive grasping using (a)-(c) different configurations of palm-finger assembly in our modular soft gripper.} (\textbf{d}) Grasping multiple objects simultaneously. (\textbf{e}) Grasping a large object by hooking onto internal features (i.e., the hole), a task not feasible for conventional grippers.}
    \label{fig:diverse_grasping_config} \vspace{-3mm}
\end{figure*}

\subsection{Novel Grasping Modalities}

Beyond adapting to a wide range of single objects, we demonstrated that the unique architecture of our soft, morphing gripper unlocks novel manipulation modalities not achievable with standard designs. Fig.~\ref{fig:diverse_grasping_config} showcases two such examples that highlight the gripper's advanced versatility can be achieved by modulating the palm-finger assembly.

Fig.~\ref{fig:diverse_grasping_config}d illustrates a stable \textbf{in-hand multi-object grasp} for in-hand multi-object assembly. Here, the gripper, set in a trapezoidal configuration, securely grasps and holds two cylindrical objects simultaneously. This is enabled by the combination of the shaped palm, which provides initial constraints, and the compliant, underactuated fingers, which can independently conform to different object surfaces within the hand. This capability is highly valuable for complex manipulation or assembly tasks that require kitting or managing multiple components at once.

Fig.~\ref{fig:diverse_grasping_config}e demonstrates a challenging \textbf{internal hook grasp} on a large object with an internal cavity. The object, a very large box with a hole inside it, is significantly larger than the gripper's maximum external grasping envelope. By lowering the gripper into the object's internal hole, the soft fingers are actuated to curl inward, forming rigid hooks that firmly engage the object's inner rim. This \textbf{grasping-from-within} strategy allows the gripper to securely lift objects that would be impossible for conventional grippers of a similar scale. 

These demonstrations confirm that the gripper's versatility extends beyond simple shape-matching, enabling complex and functionally distinct manipulation strategies.
\section{Conclusion}

This paper introduced {\textit{proprioceptive morphing}}, a paradigm for adaptive grasping that addresses the scale limitations of fixed-morphology soft grippers. By conceiving the gripper as a distributed network of modular self-sensing actuators, we enabled collaborative whole-body reconfiguration. Our experimental results validate this approach, demonstrating a dramatic expansion of the grasping envelope across a diverse range of object scales and geometries. The presented work establishes a scalable and accessible framework that offers a new pathway toward achieving biological levels of dexterity in robotic manipulation.

\vspace{1mm}
\textbf{Future Work.} Future efforts will proceed along two primary axes. First, we aim to enhance the gripper's intelligence by integrating tactile sensing and developing learning-based control policies for more complex tasks, such as in-hand manipulation. Second, we will explore the physical scalability of this modular concept, developing larger and smaller versions of the gripper for applications ranging from delicate agricultural harvesting to robust logistics and warehouse automation.





\bibliographystyle{IEEEtran}
\bibliography{main}

\end{document}